\pdfoutput=1

\documentclass[11pt]{article}

\usepackage[final]{acl}

\usepackage{times}
\usepackage{latexsym}
\usepackage{pifont}

\usepackage[T1]{fontenc}

\usepackage[utf8]{inputenc}

\usepackage{microtype}

\usepackage{inconsolata}

\usepackage{graphicx}

\usepackage{multirow}
\usepackage{booktabs}
\usepackage{comment}  
\usepackage{amsmath}
\usepackage{listings}
\usepackage{setspace}
\usepackage[hang,flushmargin]{footmisc}
\title{UCSC at SemEval-2025 Task 3: Context, Models and Prompt Optimization for Automated Hallucination Detection in LLM Output}

\author{
    \textbf{Sicong Huang} \quad \textbf{Jincheng He} \quad \textbf{Shiyuan Huang} \\
    \textbf{Karthik Raja Anandan} \quad \textbf{Arkajyoti Chakraborty} \quad \textbf{Ian Lane} \\
    University of California, Santa Cruz \\
    \texttt{\{shuan213, jhe516, shuan101, kanandan, achakr24, ialane\}@ucsc.edu}
}

\begin{document}
\maketitle
\begin{abstract}
    Hallucinations pose a significant challenge for large language models when answering knowledge-intensive queries.
    As LLMs become more widely adopted, it is crucial not only to detect \textbf{if} hallucinations occur but also to pinpoint exactly \textbf{where} in the LLM output they occur.
    SemEval 2025 Task 3, Mu-SHROOM: \textit{Multilingual Shared-task on Hallucinations and Related Observable Overgeneration Mistakes}, is a recent effort in this direction.
    This paper describes the UCSC system submission to the shared Mu-SHROOM task.
    We introduce a framework that first retrieves relevant context, next identifies false content from the answer, and finally maps them back to spans in the LLM output. The process is further enhanced by automatically optimizing prompts.
    Our system achieves the highest overall performance, ranking \#1 in average position across all languages.
    We release our code and experiment results.\footnote{\parbox{\textwidth}{\url{https://github.com/nlp-ucsc/semeval-2025-task3}}}
\end{abstract}

\section{Introduction}

Hallucinations in Large Language Model (LLM) outputs remain a significant concern \citep{sahoo-etal-2024-comprehensive, Huang_2025_hallu_survey}, undermining user trust in knowledge-intensive tasks.
In question answering, hallucinations manifest when models generate false or unverified information given world knowledge while maintaining a coherent response structure \citep{mishra2024finegrained}.

While previous research has developed metrics and benchmarks to detect the presence of hallucinations \citep{lin-etal-2022-truthfulqa, min-etal-2023-factscore}, most approaches provide only binary or scalar outputs. These measurements, though valuable, offer limited insight into the specific locations of hallucinated content, despite precise localization being crucial for fact-checking and model improvement.

The SemEval 2025 Task 3, Mu-SHROOM: \textit{Multilingual Shared-task on Hallucinations and Related Observable Overgeneration Mistakes} \citep{vazquez-etal-2025-mu-shroom}, addresses this gap by challenging participants to identify both the spans of hallucinated text and the associated confidence. The task encompasses 14 languages and evaluates system performance on Intersection-over-Union (IoU) and spearman correlation (Corr) on LLM outputs with human-annotated ground truth labels.

The UCSC team approached this challenge using a multi-step framework consisting of: \textit{(i)} context retrieval from external knowledge sources, \textit{(ii)} detection of false or unverifiable content, and \textit{(iii)} mapping error contents back to text spans. Additionally, we explored the use of automatic prompt optimization in step \textit{(ii)} and showed this further improved system performance. The proposed pipeline grounds LLM responses in the retrieved context to distinguish true from fabricated content, while prompt optimization enhances detection reliability and span labeling accuracy.

Our systems rank highly among the submitted systems, achieving a win in 5 languages and a top two position in 11 of the 14 languages on IoU and 10 of the 14 languages on Corr. Our participation in Mu-SHROOM revealed an important insight: when paired with "good context," a simple prompting-based approach can reliably detect hallucinations with better-than-human accuracy.


\begin{figure*}[]
    \centering
    \includegraphics[width=1\textwidth]{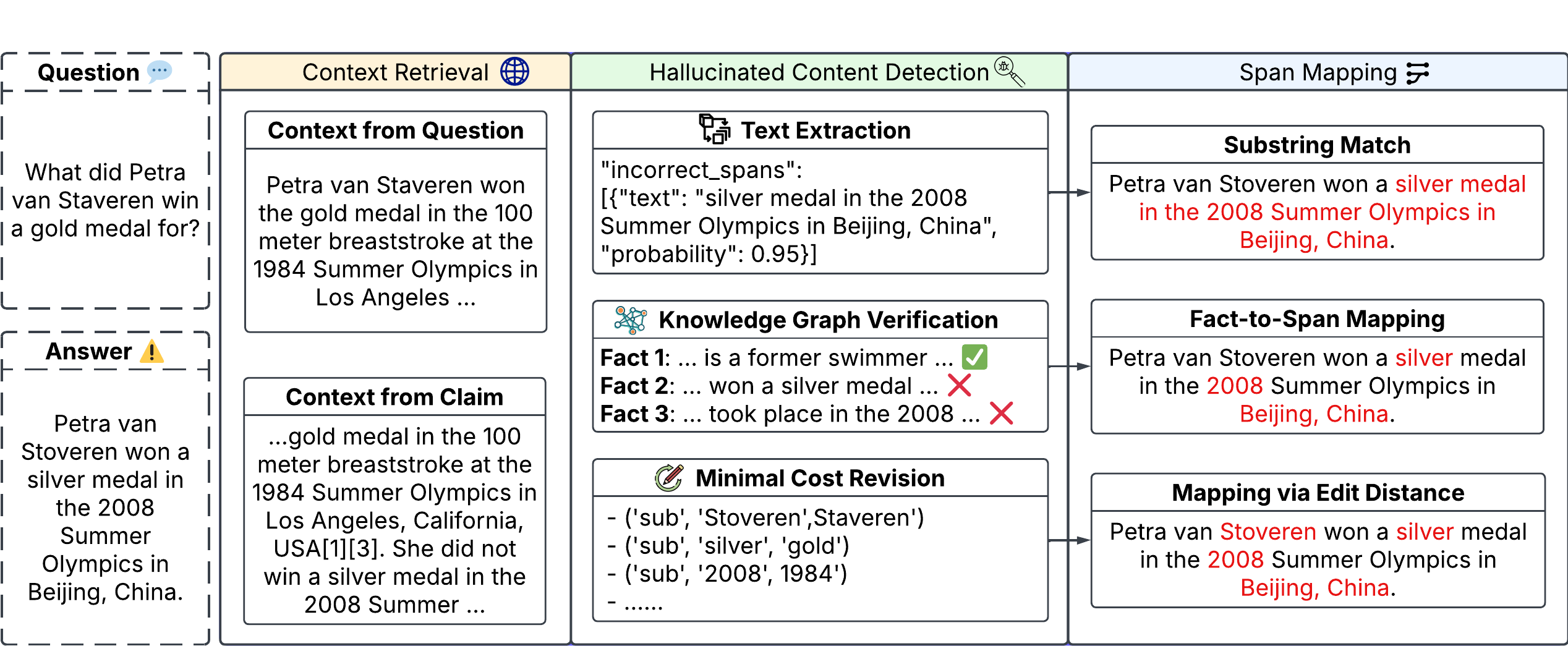}
    \caption{The UCSC hallucination detection framework. We retrieve context from external sources, identify false content in the answer, and then map these errors back to specific spans in the LLM output. In multilingual settings, we explore retrieving context either in the original language or in English by translating the question. In all cases the hallucinated content generated in the second step remains in the original language and is mapped to the answer.}
    \label{fig:sys_diagram}
\end{figure*}
\section{Background}
\subsection{Related work}
Recent efforts aiming at span-level hallucination detection for question answering, such as
HaluQuestQA \citep{sachdeva2024localizingmitigatingerrorslongform} and RAGTruth \citep{niu-etal-2024-ragtruth} provide fine-grained annotations of hallucinated spans, enabling the development of error informed refinement and retrieval-augmented fact-checking systems.
For summarization, \citet{zhou-etal-2021-detecting} proposes a token-level hallucinations prediction task and introduce a method for learning to solve the task using models fine-tuned on synthetic data.
\citet{marfurt-henderson-2022-unsupervised} proposes to detect hallucinations in an unsupervised fashion from the transformer's self-attentions. \citet{min-etal-2023-factscore} proposes to break generations down to atomic facts and assign a binary label to each fact, indicating its truthfulness.
However, despite these advances, recent benchmarks, e.g. FaithBench \citep{bao2024faithbenchdiversehallucinationbenchmark} highlight persistent challenges, as even state-of-the-art systems struggle at reliably detecting hallucinations. The SemEval-2025 Task 3 (Mu-SHROOM) builds upon these efforts by introducing a multilingual, span-level hallucination detection benchmark, pushing research toward more fine-grained, cross-lingual, and context-aware hallucination localization.

\subsection{Task Description}
The Mu-SHROOM task aims to identify hallucinated spans in LLM-generated answers across 14 languages. Human annotators provide ground truth labels: a span is a soft label if at least one annotator marks it as hallucinated and a hard label if more than half do.
Participants must predict both soft and hard label spans. Hard labels are evaluated using the character-level Intersection-of-Union metric (IoU), while soft labels are evaluated using Spearman correlation (Corr).
\section{System Overview}
\label{sec:system_overview}
Our main system adopts a three-stage pipeline (see Figure~\ref{fig:sys_diagram}), consisting of \textbf{context retrieval}, \textbf{hallucinated content detection}, and \textbf{span mapping}. 
On top of the three-stage pipeline, we use \textbf{prompt optimization} to automatically search for an optimal prompt to perform hallucinated content detection. 
In addition, we also explored a  \textbf{system combination} technique where we treated each system as an individual labeler and aggregated the results together to further increase system performance.

\subsection{Context Retrieval}
Retrieval augmented generation (RAG) \citep{lewis-2020-rag} has been shown effective at reducing hallucination at knowledge-intensive tasks \citep{jiang-etal-2023-active, gao2024retrievalaugmentedgenerationlargelanguage}.
We argue it is equally crucial to include relevant context when verifying generated text.
Step one in our pipeline is to gather information that should be helpful at either answering the input question or at confirming or refuting claims in the given answer.

\paragraph{Context from Questions} Here we use the question directly as the the search query which is passed to an external search API. The returned content is used as the context. We assume that the returned content will contain all information required to answer the input question and thus should be sufficient to verify another answer to the same question.
\newpage
\paragraph{Context from Claims} In this approach, we construct a set of queries from the claims in the answer. The resulting context can then be used to fact-check all claims in the answer, not just those directly related to the question. 
This approach will help verify claims in the answer that are missing from the context obtained from querying a search API with just the question.

\subsection{Hallucinated Content Detection}
\label{sec:hallu_detect}
In step two, we identify content in the answer unverifiable by the retrieved context.
Here we compare three distinct implementations.

\paragraph{Direct Text Extraction} We prompt the LLM to analyze the answer text and identify specific segments not verifiable from the retrieved context. The LLM compares text in the answer against the context, extracting any text spans that contain information absent from or contradicting the context.

\paragraph{Verification with Knowledge Graph} In this approach the context is parsed into a knowledge graph comprising entities and relations, while the answer is decomposed into individual facts. The LLM is then used to verify each fact by querying information about entities, checking for accuracy against the knowledge graph. This method ensures each fact is cross-verified with structured data, with the goal to enhance the reliability of the hallucination detection process.

\paragraph{Minimal Cost Revision} In this approach, we use a reasoning LLM to correct the provided answer by making the fewest possible changes. 
This method ensures that corrections are limited to only the necessary parts of the text, with the differences between the original and corrected answer being deemed hallucinated.

\subsection{Span Mapping}
After identifying the hallucinated content in the answer, we convert these broad segments into character-level spans. This conversion uses three specific methods, each corresponding to one of the three hallucinated content detection techniques:

\paragraph{Substring Match} Match the exact substring to locate the hallucinated spans within the answer. This approach is used with direct text extraction in step 2, where specific segments of text are identified as hallucinated.

\paragraph{Fact-to-Span Mapping} We prompt an LLM to map the identified false facts back to the specific spans of the answer text that generated those facts. This method is applied following verification with knowledge graph in step 2, ensuring that each false fact is accurately traced to its source text.

\paragraph{Mapping via Edit Distance} Calculate the minimum edit distance required to transform the original answer into the corrected version. During this process, all deletions and substitutions of words are identified, with these words being labeled as hallucinations. This method ensures the precise identification of unnecessary or incorrect information in the text.


\subsection{Prompt Optimization with MiPROv2}
To refine the hallucinated content detection step, we employed MiPROv2 \citep{opsahl-ong-etal-2024-optimizing}, a systematic framework for optimizing prompts in language model programs. MiPROv2 leverages Bayesian search to explore candidate prompts to optimize task metrics (e.g. IoU or Corr). In each iteration, MiPROv2 proposes updates to both the instructions and few-shot demonstrations, evaluates them on a subset of data, and uses those results to guide the next round of proposals. This process systematically discovers prompts that yield strong performance, improving the reliability of step 2.

\begin{table*}[h]
        \centering
        {
        \fontsize{8.8}{10.2}\selectfont
        \setlength{\tabcolsep}{3pt}
        \footnotesize
        \begin{tabular}{c|cc|cccccccccccccc}
            \toprule
            Model & Opt. & Trans. & ar & ca & cs & de & en & es & eu & fa & fi & fr & hi & it & sv & zh \\
            \midrule
            \multicolumn{3}{c}{} & \multicolumn{14}{|c}{IoU} \\
            \midrule
            \texttt{gpt-4o-mini} & \ding{55} & \ding{51} & 0.61 & 0.63 & 0.47 & 0.59 & 0.55 & 0.43 & 0.55 & 0.50 & 0.62 & 0.54 & 0.67 & 0.71 & 0.60 & 0.44 \\
            \texttt{gpt-4o-mini} & \ding{55} & \ding{55} & 0.59 & 0.64 & 0.48 & 0.60 & 0.56 & 0.43 & 0.57 & 0.58 & 0.60 & 0.57 & 0.68 & 0.74 & 0.63 & 0.44 \\
            \texttt{DeepSeek-R1} & \ding{55} & \ding{55}  & \textbf{\underline{0.66}} & \textbf{\underline{0.72}} & \textbf{\underline{0.54}} & \underline{0.62} & 0.57 & \textbf{\underline{0.48}} & 0.58 & 0.64 & \underline{0.63} & \textbf{\underline{0.59}} & 0.72 & 0.74 & 0.61 & 0.46 \\
            \texttt{gpt-4o}     & \ding{55} & \ding{55}  & 0.60 & 0.66 & 0.50 & 0.58 & 0.55 & 0.41 & 0.55 & 0.62 & 0.62 & \textbf{\underline{0.59}} & 0.69 & 0.72 & \underline{0.64} & 0.45 \\
            \texttt{gpt-4o}     & \ding{51} & \ding{55}  & 0.59 & 0.71 & 0.53 & 0.59 & \textbf{\underline{0.61}} & 0.40 & \textbf{\underline{0.59}} & \textbf{\underline{0.69}} & 0.62 & 0.56 & \textbf{\underline{0.74}} & \textbf{\underline{0.79}} & 0.62 & \textbf{\underline{0.47}} \\
            \midrule
            \multicolumn{3}{c|}{Multi-System Combination} & 0.65 & 0.69 & 0.53 & \textbf{0.63} & 0.58 & 0.44 & \textbf{0.59} & 0.63 & \textbf{0.65} & 0.57 & 0.71 & 0.76 & \textbf{0.65} & 0.46 \\
            \midrule
            \multicolumn{3}{c}{} & \multicolumn{14}{|c}{Corr} \\
            \midrule
            \texttt{gpt-4o-mini} & \ding{55} & \ding{51} & 0.53 & 0.71 & 0.45 & 0.57 & 0.51 & 0.53 & 0.50 & 0.54 & 0.50 & 0.47 & 0.68 & 0.70 & 0.37 & 0.29 \\
            \texttt{gpt-4o-mini} & \ding{55} & \ding{55}  & 0.52 & 0.71 & 0.50 & 0.57 & 0.51 & 0.53 & 0.51 & 0.63 & 0.48 & 0.49 & 0.72 & 0.74 & 0.33 & 0.28 \\
            \texttt{DeepSeek-R1} & \ding{55} & \ding{55}  & \underline{0.63} & \underline{0.78} & \textbf{\underline{0.58}} & \underline{0.65} & \underline{0.59} & \underline{0.60} & 0.55 & 0.68 & 0.57 & \underline{0.56} & \textbf{\underline{0.76}} & 0.77 & \underline{0.50} & 0.37 \\
            \texttt{gpt-4o}      & \ding{55} & \ding{55}  & 0.52 & 0.73 & 0.47 & 0.56 & 0.50 & 0.53 & 0.47 & 0.64 & 0.43 & 0.44 & 0.69 & 0.70 & 0.32 & 0.25 \\
            \texttt{gpt-4o}      & \ding{51} & \ding{55}  & 0.59 & 0.76 & 0.56 & 0.62 & 0.55 & 0.47 & \underline{0.58} & \textbf{\underline{0.70}} & \underline{0.58} & 0.52 & 0.76 & \textbf{\underline{0.79}} & 0.42 & \underline{0.40} \\
            \midrule
            \multicolumn{3}{c|}{Multi-System Combination} & \textbf{0.65} & \textbf{0.79} & \textbf{0.58} & \textbf{0.66} & \textbf{0.65} & \textbf{0.63} & \textbf{0.62} & 0.67 & \textbf{0.65} & \textbf{0.60} & \textbf{0.76} & \textbf{0.79} & \textbf{0.53} & \textbf{0.43} \\
            \bottomrule
        \end{tabular}
    }
    \caption{Multilingual test IoU and Corr results. \textbf{Opt.} indicates that prompt optimization was performed and \textbf{Trans.} indicates if the input was translated into English before performing context retrieval. All contexts are sourced from Perplexity Sonar Pro. IoU is used as the prompt optimization metric for all languages except English, where Corr was applied. We \underline{underline} the best-performing individual system, and \textbf{bold} the overall best.}
    \label{tab:multi_lang_test_iou}
\end{table*}

\subsection{Multilingual Systems}
Our framework design was motivated by the assumption that the pre-trained LLMs we employed might perform better in English, given the abundance of English-language training data that is generally available. For hallucination detection of non-English text use used exactly the same methods as described above. We did however explore one specific variation: we compared the use of English context vs. target-language (i.e. non-English) context for the 13 other languages within the Mu-SHROOM task.
The English-language context is obtained by translating the given questions or claims into English before retrieval. For the target-language contexts we used the question or claims in the original language to retreive the required context. In both these cases, the unverifiable content is labeled in the original language and then mapped back to the answer.

\subsection{Multi-System Combination}
\label{sec:system-combination}
Our focus in this work has been to generate hard labels for hallucinated segments with the goal to maximize IoU score. We believe this approach is best if the goal is to provide explicit feedback to users of such systems. For example when we want to highlight which segments in an LLM output could be incorrect or non-factual.
When considering the probability of a specific token in an LLM output being a hallucination or not, prior methods largely rely on the language model itself to generate a "likelihood of correctness score." Such scores are found to always be too high, as the models are overly confident of their own output. Additionally, the resulting scores do not align well with the definition of soft labels in the Mu-SHROOM task, i.e., labels based on the proportion of annotators who agree on whether certain spans are hallucinated.

In the Mu-SHROOM challenge task, we attempted to replicate the human labeling process by having multiple different systems output hard-labels. We then combined these sets of hard-labels to generate the soft-labels based on label agreement.
The expectation is that like human annotators, systems will vary in which specific tokens they label in the LLM output. For system combination in our submission systems, we combined the output of five different systems together. By treating each system as an annotator, we calculate the proportion of systems that labeled a specific span as hallucinated.

\section{Experimental Setup}

\subsection{Models and Tools}
For context retrieval, we use the \texttt{sonar-pro} model via the Perplexity API\footnote{\url{https://sonar.perplexity.ai}} (more details can be found in Appendix \ref{app:context}).
For the detection of hallucinated content, we generally use OpenAI's \texttt{GPT-4o} and \texttt{GPT-4o-mini} \citep{openai2024gpt4ocard}.
For the task of correcting answers with minimal changes, i.e. Minimal Cost Revision as described in section~\ref{sec:hallu_detect}, we found that the OpenAI \texttt{o1} reasoning model \citep{openai2024o1} out-performed the GPT-4 models. For the multilingual systems, we also evaluated the performance of \texttt{DeepSeek-R1} \citep{deepseekai2025deepseekr1incentivizingreasoningcapability} and when performing system combination, we also included Llama3.3-70B \citep{grattafiori2024llama3herdmodels} as one of the 5 systems that was combined.

We used LangChain\footnote{\url{https://langchain.com}} to build the pipeline for our submission system and to also construct the knowledge graphs\footnote{\url{https://python.langchain.com/docs/how_to/graph_constructing/}} used for the verification with knowledge-graph + fact-to-span mapping approach. DSPy \citep{khattab2024dspy} was used to perform prompt optimization. When performing prompt optimization on the validation set we perform 2-fold cross-validation to ensure reliability.

\newpage
\subsection{Annotations and Alternative Metrics}

To better understand the challenges of the hallucination span-labeling task, we manually labeled the English validation set ourselves. When we evaluated our internal annotations against the hard and soft annotations provided by the organizers, we found that our average IoU was 0.43, and the average Corr was 0.40. The best annotator in the group obtained an IoU of 0.48 and a Corr of 0.48 and the annotator with the lowest scores obtained an IoU of 0.37 and a Corr or 0.34. We found that even our best individual annotator performed significantly worse than all of our LLM-based systems. For comparison our best performing system on the validation set obtained an IoU of 0.57 and a Corr of 0.55. We hypothesize that two factors limited our overlap with the ground truth: (\textit{i}) lack of exact reference contexts, leading to discrepancies in verification, and (\textit{ii}) potential differences in labeling guidelines.

Due to the low agreement among our internal annotators, to better guide system development, we introduced a new metric MaxIoU, inspired by the maximum average Jaccard index \cite{cronin2017comparison}. MaxIoU mitigates human labeling inconsistencies by identifying the IoU with the single annotation that provides the highest IoU, rather than aggregating results into soft or hard labels. The details of the metric are provided in appendix~\ref{appendix:maxiou}.

\begin{table}[]
    \centering
    \small
    \begin{tabular}{c|c|c||c|c|c}
        \toprule
        Lang & IoU & Corr & Lang & IoU & Corr \\      
        \cmidrule{1-6}
        ar & 2 & 2 & fa & 2 & 3 \\
        ca & 1 & 1 & fi & 1 & 1 \\
        cs & 2 & 1 & fr & 5 & 3 \\
        de & 1 & 1 & hi & 2 & 2 \\
        en & 2 & 2 & it & 1 & 2 \\
        es & 6 & 1 & sv & 1 & 5 \\
        eu & 2 & 1 & zh & 9 & 9 \\
        \cmidrule{1-6}
        \multicolumn{6}{c}{Avg IoU rank: 2.6; Avg Corr rank: 2.4} \\
        \bottomrule
    \end{tabular}
    \caption{System rankings across all languages.}
    \label{tab:overall performance}
\end{table}

\section{Results \& Analyses}

\begin{table}[]
    \centering
    \fontsize{8.8}{10.2}\selectfont
    \setlength{\tabcolsep}{3pt}  
    \begin{tabular}{c|c|cc|cc}
        \toprule
        \multirow{2}{*}{Context} & \multirow{2}{*}{Method} & \multicolumn{2}{c|}{Val} & \multicolumn{2}{c}{Test} \\
        \cmidrule{3-6}
         &  & IoU & Corr & IoU & Corr \\
        \midrule
        \multirow{1}{*}{None} & Text Extr.+ Substr. Match & 0.41 & 0.45 & 0.44 & 0.43 \\
        \midrule
        \multirow{3}{*}{From Q}
        & Text Extr.+ Substr. Match & \textbf{0.55} & 0.46 & \textbf{0.56} & 0.52 \\ 
         & KG Verif.+ Fact-to-Span & 0.23 & 0.24 & 0.22 & 0.19 \\ 
         & Min-cost Revi.+ Edit Dist. & 0.52 & 0.40 & 0.53 & 0.49 \\ 
        \midrule
        \multirow{3}{*}{From C}
         & Text Extr.+ Substr. Match & 0.46 & \textbf{0.48} & 0.55 & \textbf{0.53} \\ 
         & KG Verif.+ Fact-to-Span & 0.22 & 0.24 & 0.20 & 0.14 \\ 
         & Min-cost Revi.+ Edit Dist. & \textbf{0.55} & 0.46 & 0.53 & 0.49 \\ 
        \bottomrule
    \end{tabular}
    \caption{English results of different system flows. Text extraction and knowledge graph verification use \texttt{gpt-4o-mini} and minimum cost revision uses o1.}
    \label{tab:system_flow_table}
\end{table}

\subsection{Main Results}

Across 43 participant groups, the UCSC systems consistently achieve strong performance across almost all languages. Table~\ref{tab:overall performance} shows our systems rank in the top two positions in 11 of the 14 languages on IoU and 10 of the 14 languages on Corr. Furthermore, we rank the highest in average position across all 14 languages. As our system development was focused only on English, these results demonstrate the effectiveness and generalization of our approach.

Our multilingual results are presented in Table~\ref{tab:multi_lang_test_iou}. Among single-system results with no prompt optimization or translation, DeepSeek-R1 performs the best in terms of both IoU (0.59) and Corr (0.60). However, when prompt optimization is involved, GPT-4o becomes the overall best model, although not all languages benefit from prompt optimization. Table \ref{tab:multi_lang_test_iou} also compares the effect of translating the question to English before retrieval, and the results indicate it slightly lowers performance: from 0.58 to 0.56 IoU and from 0.54 to 0.53 Corr.

Table \ref{tab:multi_lang_test_iou} further includes the best performing combined system from a diverse set of individual systems. System combination improves Corr score, by 5\% on average (between 0\% and 12\% across the 14 languages) but it generally also incurs a -5\% degradation (between 0\% and -16\% across the 14 languages) in IoU.

\subsection{Analysis of Results}
\paragraph{System Flow}
Table \ref{tab:system_flow_table} compares the performance for different system flows. We found that including retrieved context boosts performance by a considerable margin. Increasing IoU by 27\% from 0.44 to 0.56 and Corr by 23\% from 0.43 to 0.53.
Creating context from questions works slightly better than creating from claims in terms of IoU for text extraction and knowledge graph verification, but for minimum-cost revision, creating context from claims is more effective. We suspect that the reason is that the o1 model can make better use of the fact-checking information because of its reasoning abilities. The knowledge graph-based method performs significantly worse than other approaches, with IoU and Corr scores approximately 1/2 that of the other methods. Upon manual inspection, we found the knowledge graph verification step can reasonably identify false facts by querying the knowledge graph, but fact-to-span mapping is extremely unreliable, resulting in a high amount of noise in labeled spans. This results in significantly lower overall system performance. Minimum cost revision stands out as a competitive approach, although it has a significantly higher computational cost due to the reasoning required during inference.

\paragraph{Prompt Optimization}
Table~\ref{tab:prompt_optimization_comparison} shows the performance of GPT-4o with different prompt optimization targets. The performance gains from prompt optimization are evident. However, no single optimization target consistently outperforms the others across both the validation and test sets. This is likely because we use the same prompting model to propose prompts, despite optimizing different targets.

\paragraph{System Combination}
As discussed in \ref{sec:system-combination}, we explored combining predictions from multiple systems to improve correlation scores. We carefully selected a group of high-performing models that differ in architecture, context handling, and optimization strategies. We found that system combination improves Corr score, by 5\% on average (between 0\% and 12\% across the 14 languages) but it generally also incurs a -5\% degradation (between 0\% and -16\% across the 14 languages) in IoU. Details of the multilingual combination systems, including their configurations and methodologies, are provided in Appendix~\ref{appendix:multi-system}. 

\begin{table}[]
    \centering
    \footnotesize
    \setlength{\tabcolsep}{3pt}  
    \begin{tabular}{c|cc|cc}
        \toprule
        \multirow{2}{*}{Opt. Target}
        & \multicolumn{2}{c|}{Valid.}
        & \multicolumn{2}{c}{Test} \\
        \cmidrule{2-5}
        & IoU  & Corr & IoU  & Corr \\
        \midrule
        \ding{55}  & 0.44 & 0.51 & 0.55 & 0.51 \\
        IoU        & \textbf{0.57} & 0.55 & 0.60 & \textbf{0.55} \\
        Corr       & 0.54 & 0.54 & \textbf{0.61} & \textbf{0.55} \\
        MaxIoU     & 0.55 & \textbf{0.57} & 0.60 & \textbf{0.55} \\
        IoU + Corr & 0.53 & 0.56 & 0.57 & 0.53 \\
        \bottomrule
    \end{tabular}
    \caption{English results of GPT-4o on text extraction pipeline with different prompt optimization targets.}
    \label{tab:prompt_optimization_comparison}
\end{table}

\newpage
\subsection{Error Analysis}
Despite achieving remarkable performance, some limitations exist. 
The system underperforms in Chinese, likely due to the high complexity of Chinese datasets and the models' limited familiarity with this knowledge in Chinese.
Upon inspecting the system, we find that it performs well in context retrieval and hallucinated content detection, particularly through the knowledge graph verification approach. This aligns with the observations of inconsistencies in human labeling.
Moreover, our system is heavily dependent on the context and the generative labeling capabilities of the LM. Obtaining the extremely high performance of the best-performing system in this paper may not be cost effective in a real-world use case.

\section{Conclusion}
In this paper we described our system architecture, exploration and submission systems to SemEval 2025 Task 3 (Mu-SHROOM) for multilingual hallucination span labeling in LLM output. Our multistage framework, which combines context retrieval, hallucination detection, and span mapping with prompt optimization, achieves strong performance, ranking in the top two positions in 11 of the 14 languages in the evaluation set. 
Through our work, we discovered that (\textit{i}) retrieving relevant context is crucial for hallucination detection, (\textit{ii}) simple text-extraction often outperforms more complex approaches, and (\textit{iii}) prompt optimization improves system performance. 
Moreover, we find significant variations in annotated spans among human annotators, even when agreeing on underlying facts, suggesting that a more well-defined framework for annotation could benefit both automatic and human labeling of hallucinated spans.

\newpage
\bibliography{acl_latex}

\begin{thebibliography}{21}
\providecommand{\natexlab}[1]{#1}

\bibitem[{Bao et~al.(2024)Bao, Li, Qu, Luo, Wan, Tang, Fan, Tamber, Kazi,
  Sourabh, Qi, Tu, Xu, Gonzales, Mendelevitch, and
  Ahmad}]{bao2024faithbenchdiversehallucinationbenchmark}
Forrest~Sheng Bao, Miaoran Li, Renyi Qu, Ge~Luo, Erana Wan, Yujia Tang, Weisi
  Fan, Manveer~Singh Tamber, Suleman Kazi, Vivek Sourabh, Mike Qi, Ruixuan Tu,
  Chenyu Xu, Matthew Gonzales, Ofer Mendelevitch, and Amin Ahmad. 2024.
\newblock \href {https://arxiv.org/abs/2410.13210} {Faithbench: A diverse
  hallucination benchmark for summarization by modern llms}.
\newblock \emph{Preprint}, arXiv:2410.13210.

\bibitem[{Cronin et~al.(2017)Cronin, Fabbri, Denny, Rosenbloom, and
  Jackson}]{cronin2017comparison}
Robert~M Cronin, Daniel Fabbri, Joshua~C Denny, S~Trent Rosenbloom, and
  Gretchen~Purcell Jackson. 2017.
\newblock A comparison of rule-based and machine learning approaches for
  classifying patient portal messages.
\newblock \emph{International journal of medical informatics}, 105:110--120.

\bibitem[{DeepSeek-AI et~al.(2025)DeepSeek-AI, Guo, Yang, Zhang, Song, Zhang,
  Xu, Zhu, Ma, and
  Peiyi~Wang}]{deepseekai2025deepseekr1incentivizingreasoningcapability}
DeepSeek-AI, Daya Guo, Dejian Yang, Haowei Zhang, Junxiao Song, Ruoyu Zhang,
  Runxin Xu, Qihao Zhu, Shirong Ma, and et~al. Peiyi~Wang. 2025.
\newblock \href {https://arxiv.org/abs/2501.12948} {Deepseek-r1: Incentivizing
  reasoning capability in llms via reinforcement learning}.
\newblock \emph{Preprint}, arXiv:2501.12948.

\bibitem[{Gao et~al.(2024)Gao, Xiong, Gao, Jia, Pan, Bi, Dai, Sun, Wang, and
  Wang}]{gao2024retrievalaugmentedgenerationlargelanguage}
Yunfan Gao, Yun Xiong, Xinyu Gao, Kangxiang Jia, Jinliu Pan, Yuxi Bi, Yi~Dai,
  Jiawei Sun, Meng Wang, and Haofen Wang. 2024.
\newblock \href {https://arxiv.org/abs/2312.10997} {Retrieval-augmented
  generation for large language models: A survey}.
\newblock \emph{Preprint}, arXiv:2312.10997.

\bibitem[{Grattafiori et~al.(2024)Grattafiori, Dubey, Jauhri, Pandey, Kadian,
  Al-Dahle, Letman, Mathur, Schelten, and
  et~al.}]{grattafiori2024llama3herdmodels}
Aaron Grattafiori, Abhimanyu Dubey, Abhinav Jauhri, Abhinav Pandey, Abhishek
  Kadian, Ahmad Al-Dahle, Aiesha Letman, Akhil Mathur, Alan Schelten, and
  Alex~Vaughan et~al. 2024.
\newblock \href {https://arxiv.org/abs/2407.21783} {The llama 3 herd of
  models}.
\newblock \emph{Preprint}, arXiv:2407.21783.

\bibitem[{Huang et~al.(2025)Huang, Yu, Ma, Zhong, Feng, Wang, Chen, Peng, Feng,
  Qin, and Liu}]{Huang_2025_hallu_survey}
Lei Huang, Weijiang Yu, Weitao Ma, Weihong Zhong, Zhangyin Feng, Haotian Wang,
  Qianglong Chen, Weihua Peng, Xiaocheng Feng, Bing Qin, and Ting Liu. 2025.
\newblock \href {https://doi.org/10.1145/3703155} {A survey on hallucination in
  large language models: Principles, taxonomy, challenges, and open questions}.
\newblock \emph{ACM Transactions on Information Systems}, 43(2):1–55.

\bibitem[{Jiang et~al.(2023)Jiang, Xu, Gao, Sun, Liu, Dwivedi-Yu, Yang, Callan,
  and Neubig}]{jiang-etal-2023-active}
Zhengbao Jiang, Frank Xu, Luyu Gao, Zhiqing Sun, Qian Liu, Jane Dwivedi-Yu,
  Yiming Yang, Jamie Callan, and Graham Neubig. 2023.
\newblock \href {https://doi.org/10.18653/v1/2023.emnlp-main.495} {Active
  retrieval augmented generation}.
\newblock In \emph{Proceedings of the 2023 Conference on Empirical Methods in
  Natural Language Processing}, pages 7969--7992, Singapore. Association for
  Computational Linguistics.

\bibitem[{Khattab et~al.(2024)Khattab, Singhvi, Maheshwari, Zhang, Santhanam,
  A, Haq, Sharma, Joshi, Moazam, Miller, Zaharia, and Potts}]{khattab2024dspy}
Omar Khattab, Arnav Singhvi, Paridhi Maheshwari, Zhiyuan Zhang, Keshav
  Santhanam, Sri~Vardhamanan A, Saiful Haq, Ashutosh Sharma, Thomas~T. Joshi,
  Hanna Moazam, Heather Miller, Matei Zaharia, and Christopher Potts. 2024.
\newblock \href {https://openreview.net/forum?id=sY5N0zY5Od} {{DSP}y: Compiling
  declarative language model calls into state-of-the-art pipelines}.
\newblock In \emph{The Twelfth International Conference on Learning
  Representations}.

\bibitem[{Lewis et~al.(2020)Lewis, Perez, Piktus, Petroni, Karpukhin, Goyal,
  K\"{u}ttler, Lewis, Yih, Rockt\"{a}schel, Riedel, and Kiela}]{lewis-2020-rag}
Patrick Lewis, Ethan Perez, Aleksandra Piktus, Fabio Petroni, Vladimir
  Karpukhin, Naman Goyal, Heinrich K\"{u}ttler, Mike Lewis, Wen-tau Yih, Tim
  Rockt\"{a}schel, Sebastian Riedel, and Douwe Kiela. 2020.
\newblock Retrieval-augmented generation for knowledge-intensive nlp tasks.
\newblock In \emph{Proceedings of the 34th International Conference on Neural
  Information Processing Systems}, NIPS '20, Red Hook, NY, USA. Curran
  Associates Inc.

\bibitem[{Lin et~al.(2022)Lin, Hilton, and Evans}]{lin-etal-2022-truthfulqa}
Stephanie Lin, Jacob Hilton, and Owain Evans. 2022.
\newblock \href {https://doi.org/10.18653/v1/2022.acl-long.229}
  {{T}ruthful{QA}: Measuring how models mimic human falsehoods}.
\newblock In \emph{Proceedings of the 60th Annual Meeting of the Association
  for Computational Linguistics (Volume 1: Long Papers)}, pages 3214--3252,
  Dublin, Ireland. Association for Computational Linguistics.

\bibitem[{Marfurt and Henderson(2022)}]{marfurt-henderson-2022-unsupervised}
Andreas Marfurt and James Henderson. 2022.
\newblock \href {https://doi.org/10.18653/v1/2022.gem-1.21} {Unsupervised
  token-level hallucination detection from summary generation by-products}.
\newblock In \emph{Proceedings of the 2nd Workshop on Natural Language
  Generation, Evaluation, and Metrics (GEM)}, pages 248--261, Abu Dhabi, United
  Arab Emirates (Hybrid). Association for Computational Linguistics.

\bibitem[{Min et~al.(2023)Min, Krishna, Lyu, Lewis, Yih, Koh, Iyyer,
  Zettlemoyer, and Hajishirzi}]{min-etal-2023-factscore}
Sewon Min, Kalpesh Krishna, Xinxi Lyu, Mike Lewis, Wen-tau Yih, Pang Koh, Mohit
  Iyyer, Luke Zettlemoyer, and Hannaneh Hajishirzi. 2023.
\newblock \href {https://doi.org/10.18653/v1/2023.emnlp-main.741}
  {{FA}ct{S}core: Fine-grained atomic evaluation of factual precision in long
  form text generation}.
\newblock In \emph{Proceedings of the 2023 Conference on Empirical Methods in
  Natural Language Processing}, pages 12076--12100, Singapore. Association for
  Computational Linguistics.

\bibitem[{Mishra et~al.(2024)Mishra, Asai, Balachandran, Wang, Neubig,
  Tsvetkov, and Hajishirzi}]{mishra2024finegrained}
Abhika Mishra, Akari Asai, Vidhisha Balachandran, Yizhong Wang, Graham Neubig,
  Yulia Tsvetkov, and Hannaneh Hajishirzi. 2024.
\newblock \href {https://openreview.net/forum?id=dJMTn3QOWO} {Fine-grained
  hallucination detection and editing for language models}.
\newblock In \emph{First Conference on Language Modeling}.

\bibitem[{Niu et~al.(2024)Niu, Wu, Zhu, Xu, Shum, Zhong, Song, and
  Zhang}]{niu-etal-2024-ragtruth}
Cheng Niu, Yuanhao Wu, Juno Zhu, Siliang Xu, KaShun Shum, Randy Zhong, Juntong
  Song, and Tong Zhang. 2024.
\newblock \href {https://doi.org/10.18653/v1/2024.acl-long.585} {{RAGT}ruth: A
  hallucination corpus for developing trustworthy retrieval-augmented language
  models}.
\newblock In \emph{Proceedings of the 62nd Annual Meeting of the Association
  for Computational Linguistics (Volume 1: Long Papers)}, pages 10862--10878,
  Bangkok, Thailand. Association for Computational Linguistics.

\bibitem[{{OpenAI}(2024)}]{openai2024o1}
{OpenAI}. 2024.
\newblock \href {https://cdn.openai.com/o1-system-card-20241205.pdf} {Openai o1
  system card}.

\bibitem[{OpenAI et~al.(2024)OpenAI, Hurst, Lerer, Goucher, Perelman, Ramesh,
  Clark, Ostrow, Welihinda, and Alan~Hayes}]{openai2024gpt4ocard}
OpenAI, Aaron Hurst, Adam Lerer, Adam~P. Goucher, Adam Perelman, Aditya Ramesh,
  Aidan Clark, AJ~Ostrow, Akila Welihinda, and et~al. Alan~Hayes. 2024.
\newblock \href {https://arxiv.org/abs/2410.21276} {Gpt-4o system card}.
\newblock \emph{Preprint}, arXiv:2410.21276.

\bibitem[{Opsahl-Ong et~al.(2024)Opsahl-Ong, Ryan, Purtell, Broman, Potts,
  Zaharia, and Khattab}]{opsahl-ong-etal-2024-optimizing}
Krista Opsahl-Ong, Michael~J Ryan, Josh Purtell, David Broman, Christopher
  Potts, Matei Zaharia, and Omar Khattab. 2024.
\newblock \href {https://doi.org/10.18653/v1/2024.emnlp-main.525} {Optimizing
  instructions and demonstrations for multi-stage language model programs}.
\newblock In \emph{Proceedings of the 2024 Conference on Empirical Methods in
  Natural Language Processing}, pages 9340--9366, Miami, Florida, USA.
  Association for Computational Linguistics.

\bibitem[{Sachdeva et~al.(2024)Sachdeva, Song, Iyyer, and
  Gurevych}]{sachdeva2024localizingmitigatingerrorslongform}
Rachneet Sachdeva, Yixiao Song, Mohit Iyyer, and Iryna Gurevych. 2024.
\newblock \href {https://arxiv.org/abs/2407.11930} {Localizing and mitigating
  errors in long-form question answering}.
\newblock \emph{Preprint}, arXiv:2407.11930.

\bibitem[{Sahoo et~al.(2024)Sahoo, Meharia, Ghosh, Saha, Jain, and
  Chadha}]{sahoo-etal-2024-comprehensive}
Pranab Sahoo, Prabhash Meharia, Akash Ghosh, Sriparna Saha, Vinija Jain, and
  Aman Chadha. 2024.
\newblock \href {https://doi.org/10.18653/v1/2024.findings-emnlp.685} {A
  comprehensive survey of hallucination in large language, image, video and
  audio foundation models}.
\newblock In \emph{Findings of the Association for Computational Linguistics:
  EMNLP 2024}, pages 11709--11724, Miami, Florida, USA. Association for
  Computational Linguistics.

\bibitem[{V\'azquez et~al.(2025)V\'azquez, Mickus, Zosa, Vahtola, Tiedemann,
  Sinha, Segonne, S\'anchez-Vega, Raganato, Libovický, Karlgren, Ji, Helcl,
  Guillou, de~Gibert, Bengoetxea, Attieh, and
  Apidianaki}]{vazquez-etal-2025-mu-shroom}
Ra\'ul V\'azquez, Timothee Mickus, Elaine Zosa, Teemu Vahtola, J\"org
  Tiedemann, Aman Sinha, Vincent Segonne, Fernando S\'anchez-Vega, Alessandro
  Raganato, Jindřich Libovický, Jussi Karlgren, Shaoxiong Ji, Jindřich
  Helcl, Liane Guillou, Ona de~Gibert, Jaione Bengoetxea, Joseph Attieh, and
  Marianna Apidianaki. 2025.
\newblock \href {https://helsinki-nlp.github.io/shroom/} {Sem{E}val-2025 {T}ask
  3: {Mu-SHROOM}, the multilingual shared-task on hallucinations and related
  observable overgeneration mistakes}.

\bibitem[{Zhou et~al.(2021)Zhou, Neubig, Gu, Diab, Guzm{\'a}n, Zettlemoyer, and
  Ghazvininejad}]{zhou-etal-2021-detecting}
Chunting Zhou, Graham Neubig, Jiatao Gu, Mona Diab, Francisco Guzm{\'a}n, Luke
  Zettlemoyer, and Marjan Ghazvininejad. 2021.
\newblock \href {https://doi.org/10.18653/v1/2021.findings-acl.120} {Detecting
  hallucinated content in conditional neural sequence generation}.
\newblock In \emph{Findings of the Association for Computational Linguistics:
  ACL-IJCNLP 2021}, pages 1393--1404, Online. Association for Computational
  Linguistics.

\end{thebibliography}

\appendix

\label{sec:appendix}


\section{MaxIoU}
\label{appendix:maxiou}
\[
\text{MaxIoU} = \max_{i} \frac{|A \cap B_i|}{|A \cup B_i|}
\]

where \( A \) is the predicted annotation, \( B_i \) represents individual human annotations.


\section{System Combination}
\label{appendix:multi-system}
\begin{itemize}
    \item \textbf{\texttt{Llama} + Substring Match:} A system using Llama 3.3-70B with maximum substring matching,
    \item \textbf{\texttt{o1} + Minimum Edit Distance:} A system based on a reasoning model, \texttt{o1}, utilizing minimum edit distance,
    \item \textbf{Prompt Optimization Targeting IoU and Corr:} A system with prompt optimization using MiPROv2 on \texttt{gpt-4o} targeting IoU and correlation,
    \item \textbf{Prompt Optimization Targeting MaxIoU:} A system utilizing prompt optimization with MiPROv2 trained on MaxIoU in validation dataset,
    \item \textbf{\texttt{gpt-4o-mini} Reasoning}: A system using \texttt{gpt-4o-mini} to reason and map via edit distance.
\end{itemize}

\begin{table}[h]
    \centering
    \small
    \begin{tabular}{c|c|c}
        \toprule
        System & IoU & Cor \\      
        \cmidrule{1-3}
        \texttt{Llama} + Substr. Match          & 0.54                    & 0.51                     \\
        \texttt{o1} + Edit Dist.            & 0.53                    & 0.49                     \\
        Prompt Opt. (IoU \& Corr)         & 0.59                    & 0.54                     \\
        Prompt Opt. (MaxIoU)        & \underline{0.60}        & \underline{0.55}         \\
        \texttt{gpt-4o-mini} Reasoning           & 0.54                    & 0.51                     \\
        \cmidrule{1-3}
        Multi-System Combination        & \textbf{0.61}           & \textbf{0.65}            \\
        \bottomrule
    \end{tabular}
    \caption{Performance of individual systems, and the system combination. Among these systems, the combination achieves the highest IoU and significantly improves the correlation.}
    \label{tab:combination}
\end{table}

\section{Performance Evaluation By Context}
\label{app:context}
\begin{table*}[ht]
    \centering
    \fontsize{8.8}{10.2}\selectfont
    \setlength{\tabcolsep}{5pt}
    \begin{tabular}{c|c|c|c|c}
        \toprule
        Model & Context Source & Method & IoU & Corr \\
        \midrule
        \texttt{gpt-4o-mini} & \texttt{you.com} & Text Extraction + String Match & 0.5235 & \textbf{0.5270} \\ 
        \texttt{gpt-4o-mini} & \texttt{perplexity} & Text Extraction + String Match & 0.5022 & 0.4774 \\ 
        \texttt{gpt-4o-mini} & \texttt{perplexity-Llama-3.1-sonar-small} & Text Extraction + String Match & 0.5133 & 0.5058 \\ 
        \texttt{gpt-4o-mini} & \texttt{perplexity-sonar-pro} & Text Extraction + String Match & \textbf{0.5295} & 0.4554 \\ 
        \bottomrule
    \end{tabular}
    \caption{Performance evaluation by context in English in validation dataset.}
    \label{tab:performance_by_context}
\end{table*}
In Table \ref{tab:performance_by_context}, we present the performance of the text extraction system across different context sources. This evaluation is conducted on the validation dataset, focusing on the English language. We find context generated by \texttt{perplexity-sonar-pro} provides the most performance boost on IoU, thus we conduct all subsequent experiments using \texttt{perplexity-sonar-pro} context. 

\section{Prompts}
The detail of the prompt used for hallucinated content detection can be also found in the code repository.
\lstset{
  basicstyle=\ttfamily\small, 
  breaklines=true,
  breakatwhitespace=false,
  aboveskip=0pt,
  belowskip=0pt,
  lineskip=-1pt,
  xleftmargin=0pt,
  xrightmargin=0pt,
  breakindent=0pt,
}
\subsection{Text Extraction System Prompt}
\begin{lstlisting}
Based on the provided context, identify incorrect spans in the given answer text, with associated confidence levels for each incorrect portion.

You will be provided a context with a question and its corresponding answer. Your task is to identify any specific parts of the answer that describes facts that are not supported by the context. If there are multiple incorrect segments, report each one separately. Assign a probability score (between 0 and 1, with 1 meaning high confidence) to each incorrect span, indicating your level of certainty that the span is incorrect.

# Steps
1. **Read the Context**: Carefully read the provided context.
2. **Analyze the Answer**: Carefully evaluate the given answer for accuracy regarding the question and the context.
3. **Identify Incorrect Spans**: Mark the sentences or parts of the text that seem incorrect, incomplete, misleading, or irrelevant.
4. **Assign Probability**: Assign a confidence score for each answerspan you identify as incorrect:
   - A higher score indicates greater confidence that an identified segment is incorrect.
   - Provide a score for each span between 0 and 1.

# Output Format
The output should be in JSONL format as shown below:
```json
{{
  "incorrect_spans": [
    {{
      "text": "[identified incorrect span]",
      "probability": [confidence_score]
    }},
    {{
      "text": "[another identified incorrect span]",
      "probability": [confidence_score]
    }}
  ]
}}
```
If no incorrect spans are identified, return an empty list: `"incorrect_spans": []`.

# Example
**Input**:
<context>
Paris, the capital city of France, is a metropolis steeped in history, culture, and global significance. This comprehensive analysis will delve into the city's current status, basic information, and historical importance, providing a thorough understanding of why Paris is not just the capital of France, but also one of the world's most influential cities.
</context>

<question>
What is the capital of France?
</question>

<answer>
The capital of France is Berlin.
</answer>

**Output**:
```json
{{
  "incorrect_spans": [
    {{
      "text": "Berlin",
      "probability": 0.99
    }}
  ]
}}
```

# Notes
- Ensure that the probability reflects your confidence. If unsure about the degree of incorrectness, use a lower value.
- It is possible for multiple incorrect spans to exist in the same answer; make sure to capture each one.
- If the answer is fully correct, return `"incorrect_spans": []`.
- Try to identify the spans as short as possible.
- The spans should appear in the same order as they appear in the original answer.
\end{lstlisting}

\subsection{Knowledge Graph Verification System Prompt}
\begin{lstlisting}
Identify incorrect spans in the given answer text, with associated confidence levels for each incorrect portion.

You will be provided with a question and its corresponding answer. Your task is to identify any specific parts of the answer that are factually incorrect, incomplete, or misleading. If there are multiple incorrect segments, report each one separately. Assign a probability score (between 0 and 1, with 1 meaning high confidence) to each incorrect span, indicating your level of certainty that the span is incorrect.

# Steps
1. **Analyze the Answer**: Carefully evaluate the given answer for accuracy regarding the question context.
2. **Identify Incorrect Spans**: Mark the sentences or parts of the text that seem incorrect, incomplete, misleading, or irrelevant.
3. **Assign Probability**: Assign a confidence score for each span you identify as incorrect:
  - A higher score indicates greater confidence that an identified segment is incorrect.
  - Provide a score for each span between 0 and 1.

# Output Format
The output should be in JSON format as shown below:

```json
{{
    "incorrect_spans": [
    {{
        "text": "[identified incorrect span]",
        "probability": [confidence_score]
    }},
    {{
        "text": "[another identified incorrect span]",
        "probability": [confidence_score]
    }}
    ]
}}
```
- If no incorrect spans are identified, return an empty list: `"incorrect_spans": []`

# Example
**Input**:  
Question: "What is the capital of France?"  
Answer: "The capital of France is Berlin."

**Output**:
```json
{{
    "incorrect_spans": [
    {{
        "text": "Berlin",
        "probability": 0.99
    }}
    ]
}}
```

# Notes
- Ensure that the probability reflects your confidence. If unsure about the degree of incorrectness, use a lower value.
- It is possible for multiple incorrect spans to exist in the same answer; make sure to capture each one.
- If the answer is fully correct, return `"incorrect_spans": []`.
- Try to identify the spans as short as possible.
- The spans should appear in the same order as they appear in the original answer.
\end{lstlisting}
\subsection{Minimum Cost Revision System Prompt}
\begin{lstlisting}
Use the given context, correct the answer to the question with the minimum number of changes.

You will be given a context, a question and an answer to the question. The answer may not be correct. You need to make the minimum number of changes to the answer to make it correct.

Return the corrected answer wrapped in <corrected_answer> tags.

Note: Do not correct for spelling mistakes.

<context>
{context}
</context>

<question>
{question}
</question>

<answer>
{answer}
</answer>
\end{lstlisting}


\begin{table*}
    {
    \fontsize{8.2}{10.2}\selectfont
    \setlength{\tabcolsep}{3pt}
    \begin{tabular}{c|c|c|c|c|c|c|c|c|c|c|c|c}
        \toprule
        \multirow{2}{*}{Model} & \multirow{2}{*}{Context} & \multirow{2}{*}{Translation} & \multicolumn{10}{c}{IoU} \\
        & & & ar & de & en & es & fi & fr & hi & it & sv & zh \\
        \midrule
        \texttt{gpt-4o-mini} & No                   & No  & 0.5248 & 0.4359 & 0.4144 & 0.3809 & 0.4500 & 0.3841 & 0.6376 & 0.5237 & 0.5216 & 0.2722 \\
        \texttt{gpt-4o-mini} & Perplexity Sonar Pro & No  & 0.5658 & 0.5731 & \textbf{0.5503} & 0.4501 & 0.5571 & 0.5148 & 0.6277 & 0.6463 & 0.6482 & 0.3831 \\
        \texttt{gpt-4o-mini} & Perplexity Sonar Pro & Yes & 0.5968 & 0.6054 & 0.5407 & 0.4564 & 0.5434 & 0.5092 & 0.6341 & 0.6149 & 0.5870 & 0.3910 \\
        \midrule
        \texttt{DeepSeek-R1} & Perplexity Sonar Pro & No  & \textbf{0.7226} & 0.5683 & 0.4715 & \textbf{0.4828} & 0.5712 & 0.5533 & \textbf{0.7072} & \textbf{0.6975} & \textbf{0.6663} & \textbf{0.4316} \\
        \texttt{DeepSeek-R1} & Perplexity Sonar Pro & Yes & 0.6849 & 0.5480 & 0.4970 & 0.4722 & 0.5336 & 0.5064 & 0.6844 & 0.6929 & 0.6138 & 0.3859 \\
        \midrule
        \texttt{o3-mini}     & Perplexity Sonar Pro & No  & 0.5329 & 0.5944 & 0.4542 & 0.3841 & 0.4629 & 0.5443 & 0.4787 & 0.5894 & 0.5932 & 0.3988 \\
        \midrule
        \multicolumn{3}{c|}{Multi-System Combination} & 0.5862 & \textbf{0.6184} & 0.5265 & 0.4396 & \textbf{0.5813} & \textbf{0.5577} & 0.6521 & 0.6368 & 0.6481 & 0.3882 \\
        \bottomrule
    \end{tabular}
    }
    \caption{Multi-lingual validation IoU results without prompt optimization.}
    \label{tab:multi_lang_val_iou_no_prompt_opt}
\end{table*}

\begin{table*}
    \centering
    {
    \fontsize{8.8}{10.2}\selectfont
    \setlength{\tabcolsep}{3pt}
    \begin{tabular}{c|c|c|c|c|c|c|c|c|c|c|cc}
        \toprule
        \multirow{2}{*}{Model} & \multirow{2}{*}{Prompt Opt Metric} & \multicolumn{10}{c}{IoU} \\
        & & ar & de & en & es & fi & fr & hi & it & sv & zh \\
        \midrule
        \texttt{gpt-4o}      & IoU  & 0.5996 & 0.6612 & 0.5377 & 0.4197 & 0.5316 & 0.5504 & 0.6779 & 0.6701 & 0.6263 & 0.4171 \\
        \texttt{gpt-4o-mini} & IoU  & 0.5579 & 0.5710 & 0.5615 & 0.3974 & 0.4861 & 0.4930 & 0.6385 & 0.5812 & 0.5663 & 0.4271 \\
        \texttt{gpt-4o-mini} & Corr & 0.5101 & 0.5171 & 0.5314 & 0.4461 & 0.5239 & 0.5047 & 0.6208 & 0.6164 & 0.5952 & 0.3634 \\
        \bottomrule
    \end{tabular}
    }
    \caption{Multi-lingual validation IoU results with prompt optimization.}
    \label{tab:multi_lang_val_iou_with_prompt_opt}
\end{table*}

\end{document}